\documentclass{article}
\PassOptionsToPackage{table}{xcolor}
\usepackage{iclr2027_conference,times}

\usepackage{amsmath,amsfonts,bm}

\def\1{\bm{1}}

\DeclareMathAlphabet{\mathsfit}{\encodingdefault}{\sfdefault}{m}{sl}
\SetMathAlphabet{\mathsfit}{bold}{\encodingdefault}{\sfdefault}{bx}{n}

\newcommand{\Retwo}{\ensuremath{\mathrm{Re}^2}}
\newcommand{\NoRe}{No-\Retwo}
\newcommand{\TRe}{T-\Retwo}
\newcommand{\PGRe}{PG-\Retwo}
\newcommand{\VRe}{V-\Retwo}
\newcommand{\SVRe}{PV-\Retwo}

\usepackage{hyperref}
\hypersetup{hidelinks,pdftitle={Rethinking Multi-Image Re-Representation in Multi-Image Understanding}}
\usepackage{url}
\usepackage[table]{xcolor}
\usepackage{booktabs,multirow}
\usepackage{graphicx}
\usepackage{wrapfig}
\usepackage{tikz}
\usepackage[most]{tcolorbox}
\usepackage{rotating}
\newtcolorbox{promptbox}{breakable, enhanced jigsaw, colback=gray!5, colframe=gray!55, boxrule=0.4pt, arc=2pt, left=6pt, right=6pt, top=5pt, bottom=5pt, before skip=6pt, after skip=6pt}
\usepackage{amsmath,amssymb,mathtools}
\usepackage{tabularx,array}
\usepackage{xspace,microtype}
\usepackage{xurl}
\usepackage{float}
\usepackage{longtable}
\usepackage{placeins}
\usepackage{etoolbox}
\usepackage{needspace}
\usepackage{capt-of}
\usepackage{amsthm}

\theoremstyle{definition}
\newtheorem{definition}{Definition}
\newcommand{\methodname}{Mosaic\xspace}
\newcommand{\benchmarkname}{MosaicBench\xspace}
\newcommand{\trainsetname}{the training suite\xspace}
\newcommand{\modelname}{MosaicAgent-8B\xspace}

\newcommand{\numtasks}{32\xspace}

\title{Rethinking Multi-Image Re-Representation in Multi-Image Understanding}

\author{Gengyuan Zhang\thanks{Equal contribution.}\hspace{0.5em}\thanks{Corresponding author: \texttt{zhang@dbs.ifi.lmu.de}.} \\
LMU Munich \\
MCML
\And
Xiao Han\footnotemark[1] \\
LMU Munich
\And
Xinyu Xie \\
LMU Munich
\AND
Tong Liu \\
LMU Munich \\
MCML
\And
Volker Tresp \\
LMU Munich \\
MCML
}
\hypersetup{pdfauthor={Gengyuan Zhang, Xiao Han, Xinyu Xie, Tong Liu, Volker Tresp}}

\iclrfinalcopy
\begin{document}

\maketitle
\lhead{Preprint}

\begin{abstract}

Multi-image understanding requires MLLMs not only to recognise
the content of individual images, but also to organise visual
evidence distributed across them.
We study this problem through \emph{multi-image re-representation},
viewing prompted Chain-of-Thought reasoning and agentic visual tool use
as different ways of re-organising visual evidence during reasoning. We introduce \methodname{}, a general-purpose multi-image visual harness
that enables an MLLM to actively construct visual intermediates with
ten composable image operations.
We compare five re-representation settings on existing multi-image
benchmarks and on \benchmarkname{}, a new grounding-focused benchmark
for fine-grained multi-image understanding.
Our experiments show that the relative benefits of textual and visual
re-representation are strongly task-dependent.
Visual re-representation is particularly effective for tasks requiring
precise visual evidence, including hypothesis testing, precision comparison,
and orientation-sensitive reasoning, while tasks dominated by higher-level
semantic content show smaller or less consistent gains.
Building on this finding, we train \modelname{} to use \methodname{}
with reinforcement learning using only accuracy and format rewards.
Without demonstration trajectories or rewards for specific tool-use,
the agent learns to compose visual operations over multiple steps and
exhibits diverse problem-solving patterns unpromptedly.
Code and data will be released at \url{https://github.com/gengyuanmax/Mosaic}.
\end{abstract}

\section{Introduction}\label{sec:intro}
Current multimodal large language models (MLLMs) handle varied vision tasks
using contexts that interleave text and multiple images, across domains
such as general visual question answering, video understanding,
and embodied vision. Many such tasks require more than recognising the content of each
image independently: the relevant evidence may be subtle, differently oriented, distributed across images, or only apparent after the images are transformed or combined.
A conventional MLLM encodes each image into visual tokens and reasons over them together with text~\citep{bai2025qwen3,an2025llava}.
Cross-image relations can therefore be represented implicitly in the model's context.
During reasoning, however, the same visual evidence can also be
re-organised into intermediate representations that make
task-relevant details or relations easier to use.
We refer to this process as \emph{multi-image re-representation}: constructing intermediate representations from the source images to support subsequent reasoning.
This formulation also covers single-image inputs, since iterative visual operations can create multiple derived views that the model reasons over.

Re-representation can take different forms.
Multimodal chain-of-thought (CoT) organises evidence in language,
for example by describing image content, referring back to source images, and expressing relations between them~\citep{wei2022chain}.
By contrast, agentic approaches advocate revisiting the raw vision space instead to construct new image views that the model can inspect during subsequent reasoning.
Recent Thinking-with-Images methods~\citep{zheng2025deepeyes,su2025thinking,hu2024visual} enable operations such as
cropping or zooming during inference as active perception.
The distinction is simple but important: textual re-representation changes how visual evidence is expressed in language, whereas visual re-representation can also change how that evidence is presented.

\textbf{RQ1:} This raises a basic question for multi-image understanding:
\emph{when is visual re-representation more useful than textual
reasoning?}
The answer is unlikely to be uniform across tasks.
Some questions can be answered from the semantic content already
available across the images, whereas others depend on precise
appearance, orientation, spatial relations, or the visual consequences
of applying a transformation.
As we later show in Sec.~\ref{sec:existing-benchmarks}, existing multi-image
benchmarks mix these different demands, making it difficult to identify
when the visual re-representation itself is beneficial.

We study this question by comparing several forms of
multi-image re-representation within a common framework.
Our settings range from direct answering and free-form textual
reasoning to guided textual re-description, online visual
re-representation, and prefabricated visual intermediates.
To instantiate visual re-representation, we introduce
\methodname{}, a multi-image visual harness that allows an MLLM
to construct and reuse intermediate image views through operations
such as cropping, geometric transformation, and image composition.
This lets us compare not only textual and visual forms of
re-representation, but also the use of visual intermediates with their online construction.
We find that the relative benefits of textual and visual
re-representation are strongly task-dependent.
Visual re-representation is particularly effective for tasks that
require precise visual evidence or fine-grained relations across
images, while tasks dominated by higher-level semantic content
often show smaller or inconsistent gains.
To examine these cases in greater detail, we introduce
\benchmarkname{}, a grounding-focused benchmark for fine-grained multi-image understanding.

\textbf{RQ2:} These results motivate a second question:
\emph{how does an agent learn to construct useful visual
re-representations?}
Unlike textual reasoning, visual re-representation requires the agent
to choose which image assets to operate on, which operations to apply,
and how to proceed from the resulting views.
We train a model with \methodname{} using reinforcement learning with only accuracy and format rewards, without demonstration
trajectories or rewards for specific tool sequences.
We find that this is sufficient for the agent to learn multi-step compositions of visual operations.
Its trajectories also exhibit diverse problem-solving patterns that are not explicitly prescribed by the training objective.

Our main contributions include:

\begin{enumerate}
    \item formulating \emph{multi-image re-representation} and defining five re-representation settings.

    \item introducing \methodname{}, a multi-image visual harness for constructing, retaining, and reusing visual intermediates through composable image operations.

    \item introducing \benchmarkname{} and \trainsetname{}, providing grounding-focused evaluation and training data for fine-grained multi-image understanding.

    \item training \modelname{} with simple rewards and showing that it learns multi-step visual tool use with diverse problem-solving behaviours.
\end{enumerate}

\section{Related Work}\label{sec:related}
\textbf{Multi-image understanding.}
Multi-image understanding spans temporal reasoning over video frames,
spatial reasoning across views, and comparison across image
collections~\citep{meng2024mmiu}.
MANTIS develops multi-image abilities through interleaved
instruction tuning~\citep{jiang2024mantis}.
BLINK evaluates perception-intensive tasks, MMIU covers diverse
semantic, temporal, and spatial relationships, and M4Bench tests
alignment and discrimination across domains and
granularities~\citep{fu2024blink,meng2024mmiu,ye2025m4bench}.

\textbf{MLLMs for multi-image inputs.}
MLLMs accommodate multi-image inputs by conditioning language generation on interleaved images and text~\citep{alayrac2022flamingo,jiang2024mantis}.
In this \emph{encode-then-reason} setting, cross-image relationships
are established through computation over the encoded visual context.
Explicit textual reasoning provides an additional means of organising
visual evidence: Multimodal-CoT generates intermediate
rationales~\citep{zhang2023multicot}, while PromptCap and QG-CoC
use question-guided descriptions and multi-image caption chains,
respectively~\citep{hu2023promptcap,kao2025qg}.

\textbf{Multi-image reasoning with visual agents.}
Thinking with Images enables models to inspect and manipulate visual
inputs through external tools~\citep{su2025thinking}.
DeepEyes learns local image inspection, while Visual Sketchpad and
PyVision construct visual intermediates through sketching and executable
code~\citep{zheng2025deepeyes,hu2024visual,zhao2025pyvision}.
VipAct combines focused captioning and multi-image comparison agents
with perception tools~\citep{zhang2026vipact}.
Recent works focus on training such tool-mediated agents with reinforcement learning~\citep{wu2025vtoolr1,su2025openthinkimg,zhao2026pyvisionrl}.
Task-specific agents for multi-image reasoning are also an emerging research direction~\citep{wang2025imagent,wu2026reinforcing}. 

\section{Multi-Image Re-representation for Multi-Image Understanding}
\label{sec:method}

\subsection{Multi-Image Re-representation}
\label{sec:rerepresentation}

Answering a multi-image question can require grounding visual evidence dispersed across images.
Re-representation organises this evidence into intermediate records for reasoning.
Let $x=(q,I_1,\ldots,I_n)$ contain a question and its ordered source images,
and let $y$ denote an answer.

\begin{definition}[Multi-image re-representation]
\label{def:rerepresentation}
Multi-image re-representation constructs a variable-length intermediate
record $z\in\mathcal Z_m$ from $x$ to organise visual evidence for
answering $q$.
A re-representer is specified by the conditional distribution
$p_\theta^m(z\mid x)$, where
$m\in\{\mathrm{text},\mathrm{visual}\}$ denotes the representation form.
\end{definition}

A solver $p_\phi(y\mid x,z)$ uses the record alongside the original input,
giving the answer distribution
\begin{equation}
    p_{\theta,\phi}^{m}(y\mid x)
    = \sum_{z\in\mathcal Z_m}
      \underbrace{p_\theta^{m}(z\mid x)}_{\text{re-representer}}
      \underbrace{p_\phi(y\mid x,z)}_{\text{solver}},
    \qquad m\in\{\mathrm{text},\mathrm{visual}\}.
    \label{eq:rerepresentation_marginal}
\end{equation}
The re-representer and solver are usually the same model.

\begin{definition}[Textual re-representation]
\label{def:textual_rerepresentation}
Textual re-representation uses a text sequence
$z=d=(d_1,\ldots,d_L)\in\mathcal V^*$ as its intermediate record,
where $\mathcal V$ is the text vocabulary.
The sequence is generated autoregressively conditioned on $x$.
\end{definition}

The textual record can describe visual content and express cross-image
relations.
Tool interaction additionally allows the model to construct new image
views for inspection~\citep{yao2022react,hu2024visual}.

\begin{definition}[Visual re-representation]
\label{def:visual_rerepresentation}
Visual re-representation constructs visual intermediates through image
operations.
In the online setting, its record is the multimodal interaction trace
$z=\tau=(a_1,o_1,\ldots,a_T,o_T)$, where $a_t$ contains model-generated
text and a tool request or stop action, and $o_t$ is the returned
observation.
\end{definition}

The trace ends with a stop action and an empty observation.
Its distribution factorises as
\begin{equation}
p_\theta^{\mathrm{visual}}(\tau\mid x)
=
\prod_{t=1}^{T}
p_\theta(a_t\mid x,\tau_{<t})\,
P_{\mathrm{env}}(o_t\mid x,a_t,\tau_{<t}).
\label{eq:visual_trace}
\end{equation}
The environment transforms or combines source images and retained
intermediates.
Each result is returned as a rendered image with a reusable reference
and remains available for subsequent operations.
The harness and tool interface are described in
Sec.~\ref{sec:harness}.

Both forms of re-representation organise evidence from the provided input.
The following property characterises their information content relative to that input.

\textbf{Closed-evidence re-representation.}
Let $X$, $Z$, and $Y^\star$ denote the random variables corresponding
to the complete original input $x$ (question and source images before
visual tokenisation), the intermediate record $z$, and the ground-truth
answer, respectively.
In the \emph{closed-evidence} setting, the model and tools obtain no
external evidence or additional observations.
With model parameters and tool implementations fixed, write
$Z=g(X,U)$, where $U$ collects the randomness used to generate the
record and satisfies $U\perp Y^\star\mid X$.
Thus $Y^\star\perp Z\mid X$, implying
\begin{equation}
    I(Y^\star; Z \mid X) = 0.
    \label{eq:closed_evidence}
\end{equation}
This conditional-independence property follows from the stated
assumptions; it is not an empirical finding.
It concerns information about the answer beyond the complete input $X$.
Cropping and re-encoding may expose details absent from the initial
compressed visual tokens and improve their accessibility to the solver;
conditional independence need not hold when conditioning only on those tokens.

\textbf{Re-representation settings.}
We compare five settings that retain the original images and question and share the final-answer format.

\textit{A. No explicit re-representation (\NoRe{}).}
The model returns only the final answer, without an explicit
intermediate reasoning trace.

\textit{B. Free-form textual re-representation (\TRe{}).}
The model generates a free-form reasoning trace before answering,
using standard CoT prompting~\citep{wei2022chain}.

\textit{C. Prompt-guided textual re-representation (\PGRe{}).}
We extend \TRe{} with question-guided re-description instructions:
describe relevant visual content, identify its source images,
and organise cross-image comparisons~\citep{kao2025qg}.

\textit{D. Visual re-representation (\VRe{}).}
The model performs online visual re-representation with \methodname{},
following the interaction process in Eq.~\eqref{eq:visual_trace}.

\textit{E. Prefabricated visual re-representation (\SVRe{}).}
We provide visual intermediates constructed in advance alongside
the original input, omitting the interaction traces used to produce them.
The model uses the same answer-stage prompt as \TRe{} without tool access.
This setting separates the use of visual intermediates from their construction.

\begin{figure}[htbp]
    \centering
    \includegraphics[width=0.96\linewidth]{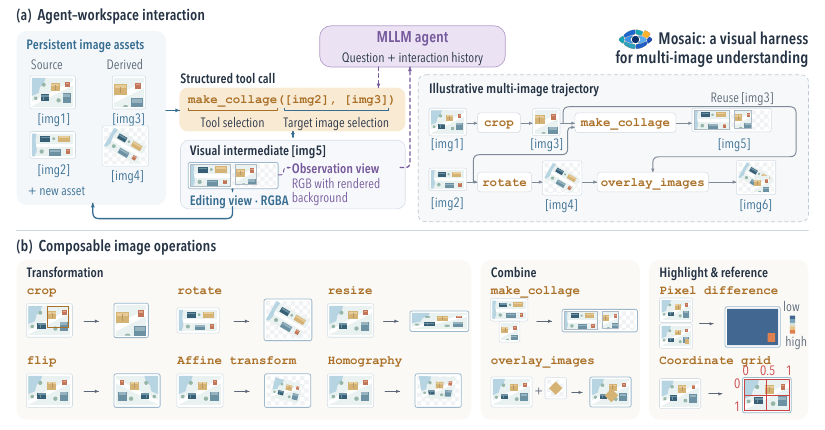}
    \vspace{-0.4cm}
    \caption{
\textbf{\methodname{}: a visual harness for multi-image understanding.}
(a) The model selects source or derived image assets, issues a
structured operation call, and inspects the returned view.
Each output is retained as a new asset for subsequent reuse.
(b) Image operations transform individual views, combine evidence
across images, and expose pixel-level differences.
}
\label{fig:mosaic_harness}
\end{figure}

\subsection{\methodname{}: A Multi-Image Visual Harness}
\label{sec:harness}
Multi-image reasoning can require selecting and re-organising visual
evidence within and across images.
\methodname{} provides a persistent workspace in which an MLLM
transforms and combines source images and retained intermediates
into task-driven visual representations
(Fig.~\ref{fig:mosaic_harness}).

\textbf{Persistent image workspace.}
Source images and derived views are stored as separate assets with
stable references, such as \texttt{[img1]}.
Operations create new assets without modifying their inputs,
allowing the model to build on a result or revisit an earlier version.
In the shared workspace, each image asset has its own transparent canvas
that expands to accommodate transformed or composited content beyond its
original boundaries.
Regions without image content remain transparent for subsequent
composition. Each asset has an editing view (RGBA image) for subsequent processing and an observation view  (RGB image) with transparent checkered background as MLLM inputs.

\textbf{Composable image operations.}
\methodname{} provides ten deterministic image operations
on the source images and their derivatives.
The model can construct visual representations over successive calls,
using the output of one operation as the input to another.
Geometric operations select or transform individual views;
collage and overlay combine content from multiple assets.
Pixel differencing exposes intensity discrepancies between aligned
images, while a coordinate grid provides spatial references.
Each operation is invoked through a structured call specifying
the input assets and parameters.
The harness stores the output as a new asset and returns its
rendered view and reference to the model.
Full tool specifications and rendering details are provided in
Appx.~\ref{app:toolbox}.

\begin{figure}[htbp]
    \centering
    \includegraphics[width=0.96\linewidth]{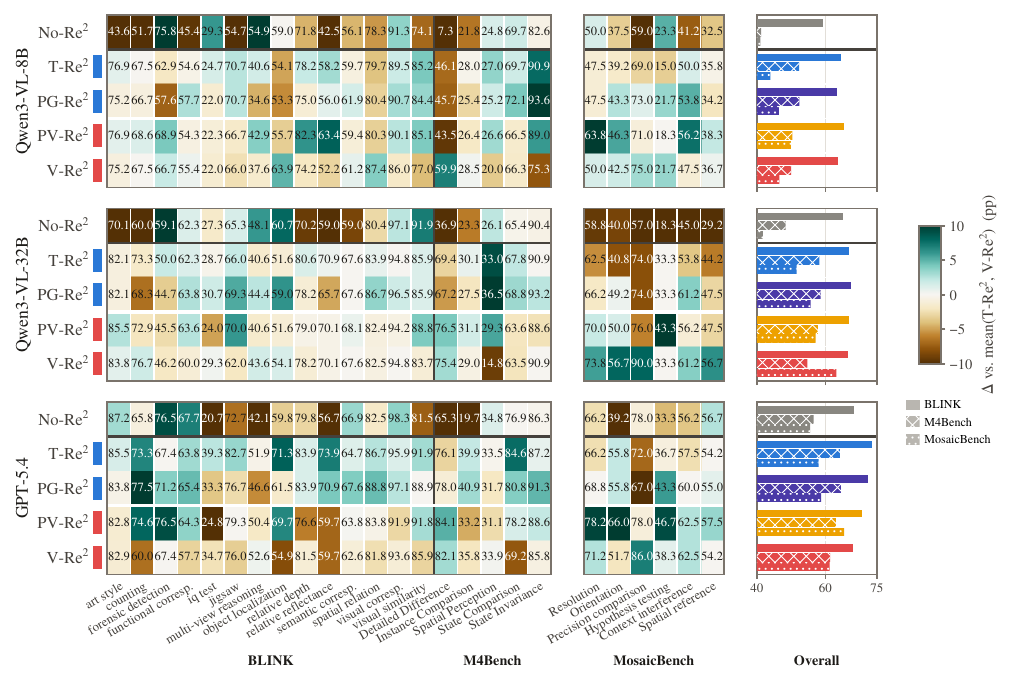}
    \vspace{-0.6cm}
\caption{
\textbf{Task-level comparison of re-representation methods.}
Rows show the five settings for Qwen3-VL-8B,
Qwen3-VL-32B, and GPT-5.4.
Cell values are accuracy (\%) on BLINK and M4Bench tasks
and on \benchmarkname{}.
from the mean of \TRe{} and \VRe{} for each model and task.
Bars on the right show overall benchmark accuracy.
}
\label{fig:rerepresentation-comparison}
\end{figure}

\section{Comparing Multi-Image Re-representation Methods}
\label{sec:representation-comparison}

We first ask which form of multi-image re-representation is beneficial for which tasks.

\subsection{Evaluating on Existing Benchmarks}
\label{sec:existing-benchmarks}

\textbf{Evaluation setup.}
We evaluate on BLINK~\citep{fu2024blink},
M4Bench~\citep{ye2025m4bench} with fine-grained subtasks.
We compare \NoRe{}, \TRe{}, \PGRe{}, \SVRe{}, and \VRe{}
for Qwen3-VL-8B, Qwen3-VL-32B, and GPT-5.4, which are representative models with basic tool-calling capabilities.
The prefabricated visual intermediates for \SVRe{} are generated in advance by
\modelname{} using \methodname{} and are shared across all solvers.

\textbf{The benefits vary across tasks.}
On M4Bench's Detailed Difference task, Qwen3-VL-8B improves
from 7.3\% under \NoRe{} to 46.1\% under \TRe{}
and 59.9\% under \VRe{}.
Prompt-guided textual re-description reaches 45.7\%.
Visual re-representation also improves Detailed Difference
over \TRe{} for Qwen3-VL-32B and GPT-5.4,
by 6.0 percentage points each.
On State Comparison, however, all three models perform
worse under \VRe{} than under \TRe{}.
The effect of re-representation therefore depends on
the subtask required, even within the same benchmark.

\textbf{Visual inputs and online construction have different effects.}
On BLINK's Relative Depth task, Qwen3-VL-8B reaches
82.3\% with prefabricated visual intermediates,
compared with 78.2\% under \TRe{} and 74.2\% under \VRe{}.
Thus, providing visual intermediates can improve performance
on a task where online re-representation does not.
This distinction motivates examining both the usefulness
of visual representations and the policy that constructs
and uses them.

These results motivate a more focused evaluation of tasks
that require fine-grained visual recognition, geometric reasoning,
and spatial localisation.
We introduce \benchmarkname{} next to examine these tasks
in greater detail.
The differences between \SVRe{} and \VRe{} motivate studying
how visual intermediates are constructed during interaction.
We next train a policy with \methodname{} and examine changes
in its performance and re-representation behaviour.

\begin{figure}
    \centering
    \includegraphics[width=0.96\linewidth]{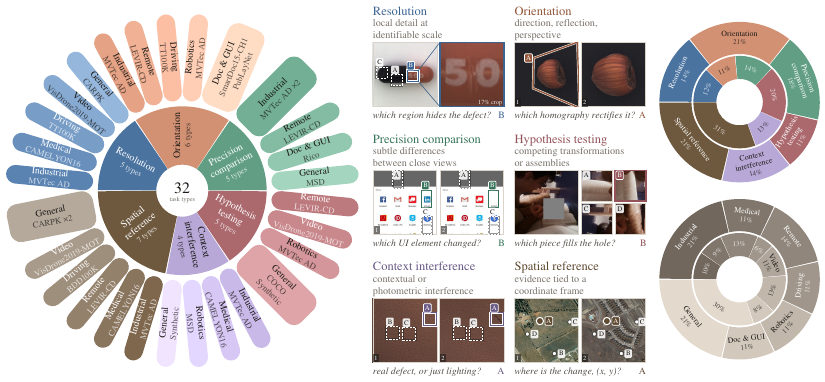}
    \vspace{-0.4cm}
    \caption{\textbf{\benchmarkname{} overview.}
    Left: task types organised by their categories and domains.
    Center: representative multi-image examples.
    Right: distributions across task (top) and domains (bottom).}
    \label{fig:data}
\end{figure}

\subsection{\benchmarkname: A Grounding-Focused Visual Understanding Benchmark}
\label{sec:mosaicbench}

We introduce \benchmarkname{} to evaluate re-representation on
tasks requiring precise visual evidence and relations.

\textbf{Task coverage.}
We construct \numtasks{} task types for training and evaluation,
grouped by their core visual challenge: resolution, orientation,
precision comparison, hypothesis testing, context interference,
and spatial reference.
Each example contains one to five images and a task-specific question.
Fig.~\ref{fig:data} illustrates representative tasks
and their visual evidence.

\textbf{Data construction.}
We draw images and annotations from 12 public datasets (VisDrone2019-MOT~\cite{Wen_2019_ICCV}, TT100K~\cite{zhu2016tt100k}, CAMELYON16~\cite{ehteshami2017diagnostic}, MVTec AD~\cite{bergmann2019mvtec}, CARPK~\cite{hsieh2017carpk}, PubLayNet~\cite{zhong2019publaynet}, LEVIR-CD~\cite{rs12101662}, SmartDoc15-CH1~\cite{burie2015smartdoc}, MSD~\cite{yang2019my}, Rico~\cite{Deka:2017:Rico}, COCO~\cite{lin2014coco}, and BDD100K~\cite{yu2020bdd100k}) to construct multi-image visual tasks from annotation augmentation.
Annotation-based selection identifies relevant objects, regions,
and correspondences.
Controlled transformations create related views through geometric
changes, local edits, and rearrangements.
Ground-truth answers follow source annotations or known
construction parameters.

\textbf{Evaluation.}
We construct \benchmarkname{} from the same generated data,
excluding overlap with \trainsetname{} at the sample, source,
and image levels.
The resulting benchmark contains 560 examples across 28 task types, with 20 examples per type.
Detailed task-construction procedures
are provided in Appx.~\ref{sec:pipeline}
and~\ref{sec:tasks_template}. Of the 28 task types, 26 use four answer options with balanced correct-answer positions.

\section{Learning Multi-Image Visual Re-representation}
\label{sec:learning}

The preceding experiments compare re-representation methods.
We now study how an agent learns to construct and use visual
intermediates.
Performance with the harness also depends on the policy's ability
to select operations and reason from their outputs.
We train the policy through reinforcement learning and analyse
changes in its performance and interaction behaviour.

\subsection{Training an Agent with \methodname{}}
\label{sec:rl}

We initialise the policy with Qwen3-VL-8B-Instruct
~\citep{bai2025qwen3} and train it using Group Relative Policy
Optimization (GRPO)~\citep{shao2024deepseekmath},
without supervised warm-up or demonstration trajectories.
We use \trainsetname{}, introduced in Sec.~\ref{sec:mosaicbench};
data selection and separation from \benchmarkname{} are detailed
in Appx.~\ref{sec:pipeline}.
Further details on the training setup are provided in Sec.~\ref{sec:training_setup}.

\textbf{Reward.}
For an interaction trace $\tau$ and final response $y$,
we assign a reward based on answer accuracy and format compliance:
\begin{equation}
    r(\tau,y,y^\star)
    =
    \mathrm{acc}(y,y^\star)
    +
    \mathrm{fmt}(\tau,y).
    \label{eq:agent_reward}
\end{equation}
Both terms are binary.
The accuracy term checks whether the extracted final answer
exactly matches the ground truth.
The format term requires a parseable answer within
\texttt{<answer>...</answer>} tags and a minimum amount
of preceding model-generated text.
For each training prompt, we sample eight rollouts and compute
advantages from group-normalised rewards.
Each episode allows up to ten assistant turns.

\textbf{Training data.}
We generate 400 examples for each of the \numtasks{} task types described in Sec.~\ref{sec:mosaicbench}, using the source datasets listed there. This yields 12,800 examples.
Using the initial policy, we perform eight tool-enabled rollouts
with \methodname{} and one no-tool run per example.
We retain examples answered incorrectly in the no-tool run
and correctly in two to seven of the eight tool-enabled rollouts.
The resulting training suite contains 3,509 examples across
32 task types.
Detailed selection protocols and overlap checks are provided
in Appx.~\ref{sec:pipeline}.

\subsection{Experimental Analysis}\label{sec:results}

\textbf{Overall performance.}
\modelname{} achieves 63.3\% on \benchmarkname{} and 61.8\%
on M4Bench, exceeding the strongest evaluated open-weight
baseline on each benchmark by 2.4 and 4.3 percentage points,
respectively (Tab.~\ref{tab:main-results}) and even on par with closed-sourced models.
On Mantis, it reaches 81.6\%, close to the highest open-weight
result of 81.9\%.
The largest advantage on \benchmarkname{} is in hypothesis
testing, where \modelname{} reaches 62.9\%, compared with
31.7\% for the strongest open-weight baseline.
It also leads the open-weight baselines in precision comparison
and orientation by 6.0 and 5.2 percentage points.
On M4Bench, the largest lead is in D.Diff, at 74.4\%
versus 66.6\%.

\begin{table*}[t]
\centering
\caption{
\textbf{Overall performance.}
We report performance on five benchmarks including \benchmarkname{}.
Bold scores mark the best reported result among the open-weight
baselines and \modelname{}.
}
\label{tab:main-results}

\small
\setlength{\tabcolsep}{2.8pt}
\renewcommand{\arraystretch}{1.12}

\resizebox{\textwidth}{!}{%
\begin{tabular}{@{}l *{7}{c} *{4}{c} ccc@{}}
\toprule

\multirow{2}{*}{\textbf{Model}}
& \multicolumn{7}{c}{\textbf{\benchmarkname{}}}
& \multicolumn{4}{c}{\textbf{M4Bench}}
& \textbf{Mantis}
& \textbf{BLINK}
& \textbf{MMIU} \\

\cmidrule(lr){2-8}
\cmidrule(lr){9-12}
\cmidrule(lr){13-13}
\cmidrule(lr){14-14}
\cmidrule(lr){15-15}

& Resol.
& Orient.
& Prec.Comp.
& Hyp.Test
& Ctx.Int.
& Spat.Ref.
& \textbf{Overall}
& D.Diff
& S.Comp
& I.Comp
& \textbf{Overall}
& \textbf{Overall}
& \textbf{Overall}
& \textbf{Overall} \\

\midrule
\multicolumn{15}{@{}l}{\textit{Open-weight baselines}} \\
\addlinespace[2pt]

InternVL3.5-8B~\citep{wang2025internvl3}
& .438 & .392 & .390 & .167 & .463 & .292 & .363
& .218 & .649 & .228 & .373
& .696 & .581 & .524 \\
\addlinespace[2pt]

LLaVA-OV-1.5-8B~\citep{an2025llava}
& .338 & .375 & .340 & .267 & .400 & .233 & .325
& .000 & .601 & .254 & .315
& .567 & .460 & .417 \\
\addlinespace[2pt]

GLM-4.6V-Flash~\citep{glm2025glmv}
& .625 & .475 & .690 & .300 & .600 & .342 & .505
& .606 & \textbf{.740} & .197 & .536
& .737 & .695 & .627 \\
\addlinespace[2pt]

MiniCPM-V-4.5~\citep{yao2025minicpm}
& .500 & .475 & .550 & .267 & .500 & .425 & .463
& .349 & .664 & .254 & .466
& .728 & .612 & .552 \\
\addlinespace[2pt]

Qwen3-VL-8B-Thinking~\citep{bai2025qwen3}
& .538 & .408 & .500 & .317 & .500 & .358 & .436
& .591 & .697 & .259 & .551
& .774 & .634 & .606 \\
\addlinespace[2pt]

Qwen3.5-27B~\citep{qwen3.5}
& \textbf{.800} & .508 & .810 & .283 & .600
& \textbf{.583} & .609
& .666 & .692 & .259 & .575
& .807 & \textbf{.723} & \textbf{.695} \\
\addlinespace[2pt]

Qwen3.5-9B~\citep{qwen3.5}
& .763 & .517 & .800 & .317 & \textbf{.688} & .525 & .607
& .657 & .659 & \textbf{.316} & .571
& .793 & .705 & .668 \\
\addlinespace[2pt]

Qwen3-VL-8B~\citep{bai2025qwen3}
& .497{\scriptsize$\pm$.005} & .429{\scriptsize$\pm$.014} & .621{\scriptsize$\pm$.026} & .179{\scriptsize$\pm$.038} & .541{\scriptsize$\pm$.058} & .381{\scriptsize$\pm$.037}
& .452{\scriptsize$\pm$.043}
& .455{\scriptsize$\pm$.024}
& .692{\scriptsize$\pm$.013}
& .271{\scriptsize$\pm$.030}
& .513{\scriptsize$\pm$.012}
& \textbf{.819}{\scriptsize$\pm$.010}
& .644{\scriptsize$\pm$.004}
& .610 \\

\midrule
\multicolumn{15}{@{}l}{\textit{Proprietary models}} \\
\addlinespace[2pt]

GPT-4o-mini~\cite{openai_gpt4o-mini}
& .388 & .425 & .320 & .183 & .363 & .233 & .325
& .004 & .654 & .311 & .334
& .724 & .566 & --- \\
\addlinespace[2pt]

GPT-5.4~\cite{openai_gpt5-4}
& .663 & .558 & .720 & .367 & .575 & .542 & .580
& .761 & .846 & .399 & .665
& .765 & .739 & --- \\
\addlinespace[2pt]

Claude Sonnet 5~\citep{anthropic_sonnet5}
& .688 & .608 & .830 & .383 & .650 & .583 & .636
& .703 & .736 & .316 & .636
& .816 & .695 & --- \\

\midrule
\multicolumn{15}{@{}l}{\textit{Visual Harness}} \\
\addlinespace[2pt]

GPT-4o-mini + \methodname{}
& .550 & .392 & .670 & .250 & .338 & .317 & .425
& .381 & .625 & .301 & .422
& .728 & .569 & --- \\
\addlinespace[2pt]

GPT-5.4 + \methodname{}
& .713 & .517 & .860 & .383 & .625 & .542 & .613
& .821 & .692 & .358 & .654
& .779 & .683 & --- \\
\addlinespace[2pt]

\textbf{\modelname{}}
& .697{\scriptsize$\pm$.026}
& \textbf{.569}{\scriptsize$\pm$.012}
& \textbf{.870}{\scriptsize$\pm$.007}
& \textbf{.629}{\scriptsize$\pm$.022}
& .575{\scriptsize$\pm$.023}
& .496{\scriptsize$\pm$.042}
& \textbf{.633}{\scriptsize$\pm$.011}
& \textbf{.744}{\scriptsize$\pm$.014}
& .704{\scriptsize$\pm$.010}
& .280{\scriptsize$\pm$.031}
& \textbf{.618}{\scriptsize$\pm$.014}
& .816{\scriptsize$\pm$.012}
& .628{\scriptsize$\pm$.010}
& .572 \\

\bottomrule
\end{tabular}%
}

\end{table*}

\setlength{\columnsep}{8pt}
\setlength{\intextsep}{8pt}
\begin{wraptable}{r}{0.49\textwidth}
    \centering
    \vspace{-1\baselineskip}

    \caption{
    \textbf{RL training with and without the visual harness.}
    Both RL policies use the same training suite.
    The backbone and harness-trained \modelname{} are evaluated
    under both \TRe{} and \VRe{}, while the CoT RL control
    is evaluated under \TRe{}.
    Results are accuracy (\%).
    }
    \label{tab:rl-ablation}

    \tiny
    \setlength{\tabcolsep}{3.5pt}
    \renewcommand{\arraystretch}{1.10}

    \resizebox{\linewidth}{!}{%
    \begin{tabular}{@{}lccccc@{}}
    \toprule
    \textbf{Training}
    & \textbf{MosaicBench}
    & \textbf{Mantis}
    & \textbf{BLINK}
    & \textbf{M4Bench}
    & \textbf{MMIU} \\
    \midrule

    \methodname{} w/o RL
    & 45.2{\tiny$\pm$1.44}
    & 81.9{\tiny$\pm$1.00}
    & 64.4{\tiny$\pm$0.40}
    & 51.3{\tiny$\pm$1.20}
    & 61.0 \\

    \midrule

    RL w/o \methodname{}
    & 53.0{\tiny$\pm$0.54}
    & 80.4{\tiny$\pm$1.47}
    & 58.2{\tiny$\pm$0.60}
    & 51.6{\tiny$\pm$0.80}
    & 56.8 \\

    \midrule

    RL w/ \methodname{}
    & 63.3{\tiny$\pm$1.10}
    & 81.6{\tiny$\pm$1.20}
    & 62.8{\tiny$\pm$1.00}
    & 61.8{\tiny$\pm$1.40}
    & 57.2 \\

    \bottomrule
    \end{tabular}
    }

\end{wraptable}

\textbf{Training dynamics and tool-use distribution.}
Over 218 RL steps, the mean accuracy reward increases from
0.48 to 0.63 and the format reward from 0.91 to 0.99,
comparing the first and last ten steps
(Fig.~\ref{fig:reward_tooluse}).
On \benchmarkname{}, total tool calls increase from 1,830
before training to 3,037 afterwards, a $1.66\times$ increase.
The distribution also shifts: cropping rises from 36.7\%
to 54.8\% of all calls, and collage from 0.5\% to 6.2\%.
Pixel differencing, by contrast, falls from 16.1\% to 3.3\%.
The trained policy thus allocates a larger share of its calls
to extracting image regions and composing views,
alongside the increase in overall tool use.

\textbf{Training with and without the harness.}
We compare the pre-RL backbone with \modelname{} and a CoT policy
trained on the same training suite without visual harness
(Tab.~\ref{tab:rl-ablation}).
RL training without harness (CoT RL) increases accuracy on \benchmarkname{} from 45.2\%
to 53.0\%, while M4Bench changes from 51.3\% to 51.6\%.
With \methodname{} available during both training and inference,
\modelname{} reaches 63.3\% and 61.8\%, exceeding the CoT RL
control by 10.3 and 10.2 percentage points, respectively.
Thus, the performance gains are not attributable to RL alone;
harness-enabled visual re-representation provides a substantial
additional benefit beyond CoT RL on the same training data.

\subsection{Does Training Improve Re-representation Construction or Utilisation?}
\label{sec:construction-utilization}

\textbf{Fixed-prefix forced-answer probing.}
To measure what can be answered from an intermediate trajectory state,
we reconstruct each recorded trajectory by deterministically replaying
its tool calls in the same visual workspace.
For question $i$, let $H^p_{i,t}$ denote the reconstructed context
after $t$ tool transitions of a trajectory produced by checkpoint $p$.
The context contains the original question and source images together
with all intermediate model text, tool calls, tool feedback, and
visual observations available up to that point.
At $t=0$, it contains only the original input.

At each prefix, we freeze the trajectory: the solver cannot continue
reasoning or invoke additional tools.
Following early-answering interventions
~\citep{lanham2023faithfulness}, we append the standard answer
instruction and an \verb|<answer>| prefix, then read the next-token
logits for the valid candidate answers.
For solver checkpoint $s$, we have
\begin{equation}
v_{p,s}(i,t)
=
\mathbb{I}\!\left[
\arg\max_{y\in\mathcal{Y}_i}
\ell_s(y\mid H^p_{i,t})
=
y_i^\star
\right].
\label{eq:cross_prefix_correctness}
\end{equation}
\emph{Prefix answerability} is the mean of $v_{p,s}(i,t)$ over trajectories.
Because probing does not alter or extend the recorded trajectory,
changes in answerability reflect the information available at each
prefix rather than additional inference performed by the solver.

We cross trajectories produced by the backbone (pre-RL) and \modelname{} (post-RL)
with both solver checkpoints.
Holding the solver fixed compares trajectory construction;
holding the trajectory fixed compares answer readout from
the same context.
We retain trajectories with at least one tool call and a valid
final response, and pair re-representer comparisons over the 441
questions meeting these criteria for both checkpoints.

\begin{figure*}[t]
    \centering
        \begin{minipage}[t]{0.57\textwidth}
        \centering
        \includegraphics[
            width=\linewidth,
            height=5.2cm,
            keepaspectratio
        ]{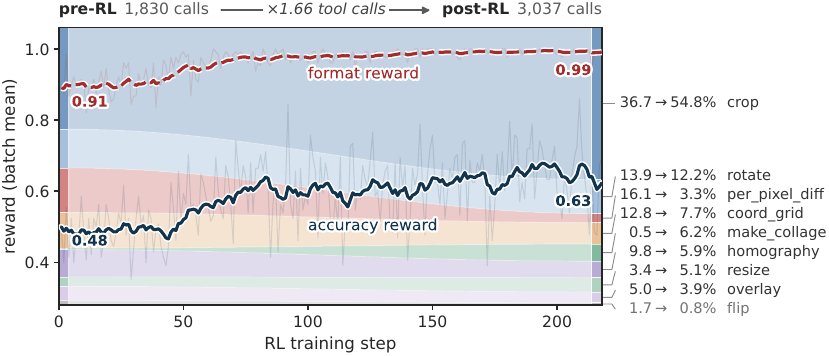}
        \vspace{-0cm}
        \captionof{figure}{
        \textbf{Training rewards and tool use shift.}
        Curves show batch-mean accuracy and format rewards over training,
        with exponential moving-average smoothing ($\alpha=0.12$).
        Background bands show each tool's usage distribution on
        \benchmarkname{} before and after training; intermediate band
        widths are interpolated for display.
        }
        \label{fig:reward_tooluse}
    \end{minipage}
    \hfill
    \begin{minipage}[t]{0.42\textwidth}
        \centering
        \includegraphics[
            width=\linewidth,
            height=5.2cm,
            keepaspectratio
        ]{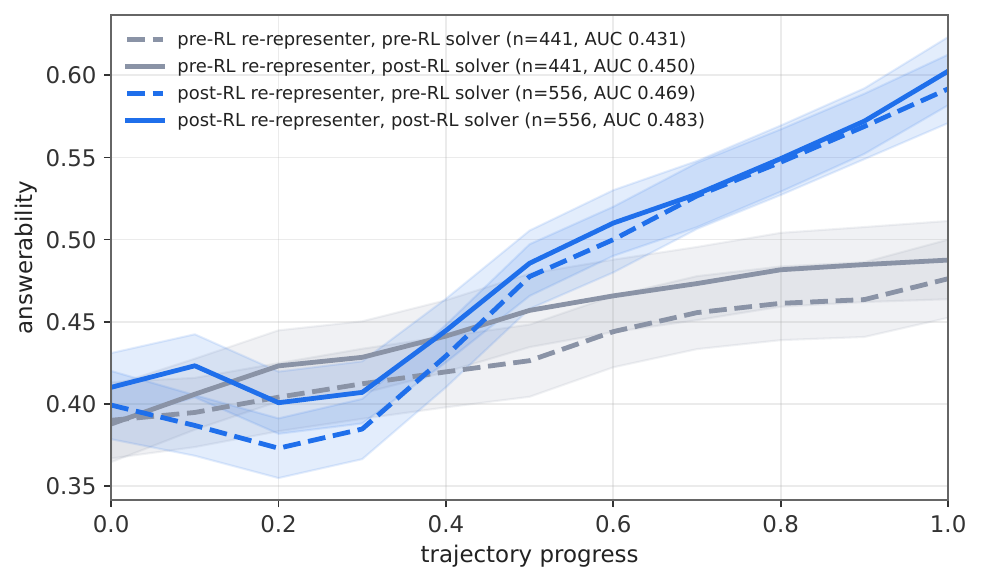}
        \vspace{-0cm}
        \captionof{figure}{
        \textbf{Crossed evaluation of trajectory construction and answer readout.}
        Colour identifies the re-representer that generated the trajectory, while line style identifies
        the solver used for forced answering.
        Bands show $\pm1$ standard error.
        }
        \label{fig:cross_answerability}
    \end{minipage}

\end{figure*}

\textbf{Post-training trajectories improve final-prefix accuracy.}
With the solver fixed, trajectories produced after training
increase final-prefix answerability by 11.0 percentage points
with the backbone solver and 10.5 points with the trained solver as shown in Fig.~\ref{fig:cross_answerability}.
For fixed trajectories, switching to the trained solver changes
the area under the normalised-progress curve by
$+0.019$ on pre-training trajectories
and $+0.014$ on post-training trajectories.
The re-representer gains transfer to both solvers, supporting improved trajectory construction as the training benefit.
\textbf{The gains emerge later in the interaction.}
Post-training trajectories contain more tool steps on average (5.41 \textit{vs.} 3.16).
We therefore also compare answerability against absolute
tool-step budgets.
The re-representer advantage is absent over the first two steps.

\subsection{How Answerability Evolves During Re-representation}
\label{sec:trajectory-profiles}
We next inquire how agents with \methodname{} proactively solve the multi-image tasks.

\begin{figure}[htbp]
    \centering
    \includegraphics[width=1\linewidth]{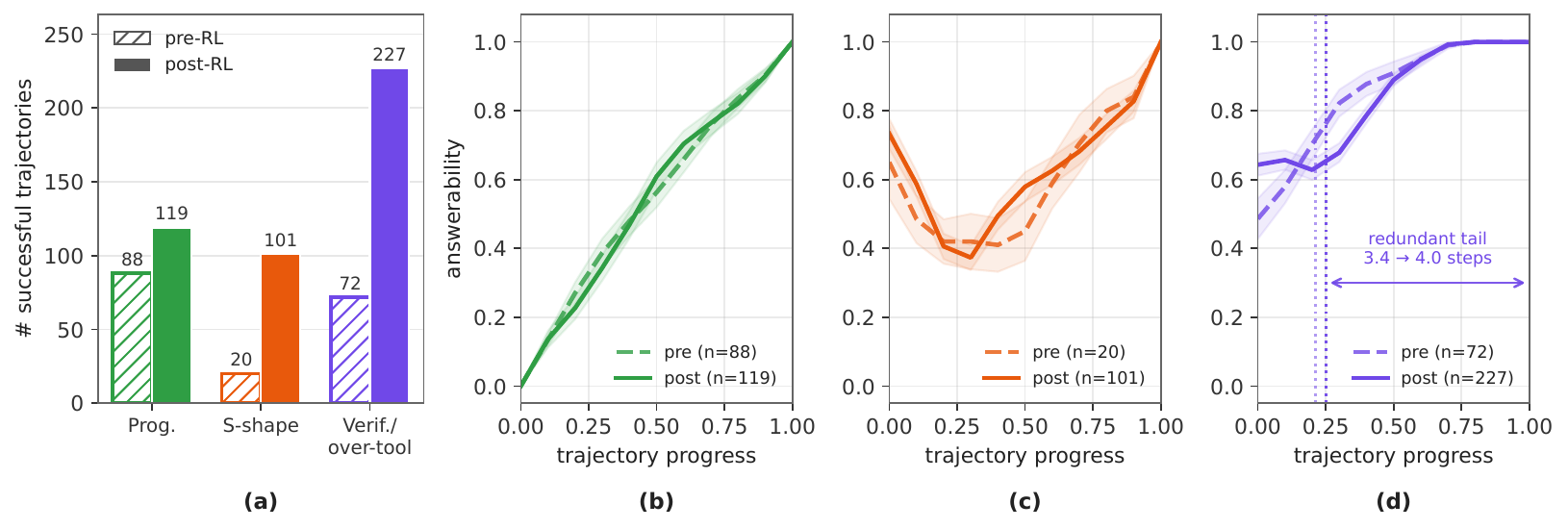}
    \vspace{-20pt}
\caption{
\textbf{Profiling answerability throughout trajectories.}
(a) Counts exhibiting progression, self-correction,
or overtooling.
(b)--(d) Mean prefix answerability within each group of trajectories on normalised trajectory progress.
Dashed lines for pre-RL and solid lines for
post-RL; bands show $\pm1$ standard error.
Dotted lines in (d) mark the mean stabilisation point where tool benefits saturate.
}
\label{fig:voi_patterns3_v2_thr2_1}
\end{figure}

\textbf{Answerability profiles.}
Among trajectories with a correct final-prefix probe, we identify
two transition patterns.
\emph{Progression} starts incorrect and becomes correct without
a later reversal, while \emph{self-correction} contains at least one
correct-to-incorrect transition followed by recovery.
Trajectories that remain correct at every prefix are
\emph{consistently correct}.
Separately, we measure \emph{post-stabilisation continuation}:
at least two additional tool steps after the probed answer stabilises,
which may reflect verification or unnecessary tool use.
Fig.~\ref{fig:voi_patterns3_v2_thr2_1} compares 210 final-prefix-correct trajectories before training and 335 after training.

\textbf{Progression and self-correction.}
Progression captures the direct case in which successive visual
operations make the available evidence sufficient for answering.
Self-correction reveals a less monotonic process: an intermediate
representation can move the model away from a correct answer before
later operations recover it.
Its increased prevalence after training suggests that successful visual
reasoning can involve revising intermediate rather than
only accumulating evidence.

\textbf{Verification and overtooling.}
Post-stabilisation continuation captures a different behaviour:
the agent keeps processing visual evidence after the probed answer has
already stabilised.
Such steps may verify an existing answer by inspecting additional
evidence, or may constitute unnecessary tool use.
Their increased frequency after training shows that the learned policy
does not simply stop once a correct answer becomes available.

\textbf{Training changes the mixture of problem-solving modes.}
Training does not simply produce more monotonic progression:
although its absolute count increases, its share decreases from
41.9\% to 35.5\%, while self-correction and post-stabilisation
continuation become substantially more frequent.
The learned policy therefore exhibits a different mixture of
progression, revision, and continued processing rather than
converging to a single strategy.

\section{Conclusion}
\label{sec:conclusion}
In this work, we studied when textual and visual re-representation
improve multi-image understanding and how an agent learns to
construct task-driven visual representations.
We introduced \methodname{}, a visual harness for re-organising
evidence within and across images, and \benchmarkname{},
a grounding-focused benchmark for multi-image understanding.
Our empirical study shows that the relative benefits of textual
and visual re-representation depend on the task.
The visual harness is particularly effective on tasks requiring
precise visual evidence, including hypothesis testing,
precision comparison, and orientation-sensitive tasks.
We further show that reinforcement learning with accuracy
and format rewards enables an agent to compose visual tools
over multiple steps.
The resulting policy exhibits diverse problem-solving patterns
without demonstration trajectories or rewards that prescribe
specific behaviours.
These findings highlight the construction of
task-driven visual representations as a learnable component
of multi-image reasoning.

\subsection*{AI use statement}
In this work, we used generative AI tools to assist with translation, provide feedback on research methodology, and support qualitative and thematic data analysis.preprint We did not use generative AI to generate dataset examples or develop theoretical models. Additionally, we used generative AI tools to create artefacts, identify relevant literature, summarise or analyse existing literature, and create or modify scientific figures or images. We reviewed all AI-assisted work. All code was reviewed by at least three authors. The authors verified all scientific claims. AI-assisted figure generation was limited to visual styling and did not modify the underlying data. We take responsibility for the final content of this work,
including text, claims or artefacts produced with the aid of generative AI.


\bibliography{references}
\bibliographystyle{references}

\appendix

\clearpage
\pretocmd{\section}{\ifinner\else\Needspace{8\baselineskip}\fi}{}{}
\pretocmd{\subsection}{\ifinner\else\Needspace{7\baselineskip}\fi}{}{}
\pretocmd{\subsubsection}{\ifinner\else\Needspace{6\baselineskip}\fi}{}{}
\section*{Appendix}
The appendix is organised as follows.
\begin{itemize}
    \item \textbf{Appendix~\ref{app:visual-harness}: \methodname{} visual harness.}
    Tool definitions, prompting setup, and training details.
    \item \textbf{Appendix~\ref{app:benchmark-details}: \benchmarkname{} dataset.}
    Task coverage, data sources and provenance, sample format,
    validation and shortcut checks, training and evaluation set construction,
    task-specific construction, and reproducibility details.
    \item \textbf{Appendix~\ref{app:extended-results}: Extended results.}
    Tool ablations beyond cropping, model comparisons, and failure-pattern analysis.
\end{itemize}

\section{\methodname{}: A Visual Harness for Multi-Image Understanding}
\label{app:visual-harness}
\methodname{} is the image workspace and executes the visual operations used in this study.
This section documents its tool interfaces, system prompts, and training setup.

\subsection{Tool Definitions}\label{app:toolbox}
The harness exposes ten composable operations over image assets.
The definitions below specify the inputs and outputs of each operation.

\paragraph{\texttt{crop}}
Crops a rectangular region from one image. It accepts an input target image and the crop box corner coordinates x1, y1, x2, y2 as inputs. It returns: a new image containing only the cropped region.

\paragraph{\texttt{rotate}}
Rotates one image by an arbitrary angle in degrees. It accepts an input target image and the angle as inputs.
It returns: a new image rotated by the angle.

\paragraph{\texttt{resize}}
Resizes one image to an exact width and height in pixels. It accepts an input target image and the width and height as inputs.
It returns: a new image at exactly width x height.

\paragraph{\texttt{flip}}
Flips one image horizontally or vertically. It accepts an input target image and the mode as inputs.
It returns: a new image flipped by the given mode.

\paragraph{\texttt{apply\_affine\_transformation}}
Applies a 2D affine transform --- rotation, uniform scaling, and translation --- to one image. It accepts an input target image and the rotation\_angle, translation\_dx, translation\_dy, and scale as inputs.
It returns: a new image with the affine transform applied.

\paragraph{\texttt{draw\_normalized\_coordinate\_grid}}
Draws a normalised coordinate grid over one image as a spatial reference. It accepts an input target image and the grid row and column numbers as inputs.
It returns: a new image with red gridlines drawn over the input and normalised 0-1 tick labels along the top and left edges.

\paragraph{\texttt{compare\_per\_pixel\_image\_difference}}
Compares two images pixel by pixel. It accepts two input images of the same size as inputs.
It returns: a heatmap image where warmer colors mark larger per-pixel differences and cool colors mark identical or near-identical pixels.

\paragraph{\texttt{apply\_homography\_transformation}}
Applies a 3x3 projective homography transform to one image. It accepts an input target image and a 3x3 homography matrix as inputs.
It returns: a new image with the homography applied.

\paragraph{\texttt{make\_collage}}
Tiles N input images into a single output image arranged as a \texttt{rows} x \texttt{cols} grid, filled in row-major order. It accepts the input target images and the rows and cols as inputs.
It returns: a new image with the input images arranged in a rows x cols grid.

\paragraph{\texttt{overlay\_images}}
Overlays the foreground image onto the background image at a normalised position. It accepts the background and foreground images and the x and y coordinates in the background image as inputs.
It returns: a new image with the foreground image composited onto the background image.

\subsection{Prompt Engineering}
We use different system prompts depending on whether visual tools are available.
The templates below document both settings and their differences.

\paragraph{Tool mode}

The system message sent when the agent has the toolbox available. The
tool signatures inside \texttt{\textless{}tools\textgreater{}} are
generated from the tool registry (one JSON object per tool, using each
tool's \texttt{description\_baseline}).

\begin{promptbox}
\ttfamily\footnotesize\raggedright\setlength{\parindent}{0pt}
\setlength{\parskip}{0pt}
You are an agent for multi-image visual reasoning in an environment that provides a suite of tools to support reasoning and task-solving through information processing, visual evidence gathering, transformation, comparison, and verification. Keep working on the user's question until it is fully resolved, and only end your turn once you are sure of the answer. The question always has a correct answer, so never give up or claim it cannot be determined --- keep reasoning and using tools until you find it.
\par
\medskip
Solve the problem step by step. Look closely at the actual content of the images and reason about what they show. When the answer cannot be determined from the images alone, use the provided tools to gather more visual evidence --- for example zooming in by cropping a region, rotating or flipping an image, comparing two images, or warping one image by a candidate transform.
\par
\medskip
You MUST use a tool WHENEVER it can improve your understanding of the images to solve the task. You MUST plan extensively before each tool call, and reflect extensively on the tool result of the previous step.
\par
\medskip
The provided images are labelled in order as [img1], [img2], ..., each followed by its pixel size --- e.g. [img1] (640x480). To pick which image(s) a tool operates on, pass only the imgN token (never the size) in that tool's image argument(s), as declared in its schema. Any image a tool produces becomes the next [imgN], labelled the same way.
\par
\medskip
Environment:
\par
Image tools operate in a shared workspace in which each image asset has its own transparent canvas. They let you transform and combine images --- including rotating, flipping, cropping, collaging (arranging images into a grid), compositing, and overlaying one image onto another. Each tool call renders its result as a new image, appended as the next [imgN], which you can inspect and operate on further. Each asset's canvas expands automatically to fit results that don't fill a plain rectangle, leaving transparent padding in the empty areas (see Conventions).
\par
\medskip
Conventions:
\par
- Coordinates: the origin (0, 0) is at the top-left corner of the rendered canvas --- the full image you see, transparent checkerboard regions included --- NOT the top-left of the opaque content within it. Coordinates (points and boxes) refer to that canvas; all tools except apply\_affine\_transformation (whose translation\_dx / translation\_dy are in pixels) take them normalised to 0-1 of the canvas size. The x-axis increases to the right and the y-axis increases downward.
\par
- Rotation: angles are measured in degrees. A positive angle rotates the image counter-clockwise; a negative angle rotates it clockwise.
\par
- Transparency: the canvas auto-expands only when an operation produces content that does not fill a plain rectangle --- e.g. rotating an image (its corners no longer align to the frame), collaging images of differing sizes, or overlaying/compositing with an offset. The resulting empty areas are shown as a gray-and-white checkerboard (the standard Photoshop "transparency" indicator). Operations whose result is already a full rectangle --- such as flipping or cropping --- produce no transparent region, and the canvas matches the image content exactly. The checkerboard is NEVER part of the image content: it only marks empty pixels and does not exist in the real image. Ignore it and reason only about the actual opaque content.
\par
\medskip
\# Tools
\par
\medskip
You may call one or more functions to assist with the user query.
\par
\medskip
You are provided with function signatures within \textless{}tools\textgreater{}\textless{}/tools\textgreater{} XML tags:
\par
\textless{}tools\textgreater{}
\par
\textit{[One JSON signature per tool (10 tools total); full descriptions omitted here for brevity.]}
\par
\textless{}/tools\textgreater{}
\par
\medskip
For each function call, return a JSON object with function name and arguments within \textless{}tool\_call\textgreater{}\textless{}/tool\_call\textgreater{} XML tags:
\par
\textless{}tool\_call\textgreater{}
\par
\{"name": \textless{}function-name\textgreater{}, "arguments": \textless{}args-json-object\textgreater{}\}
\par
\textless{}/tool\_call\textgreater{}
\par
\end{promptbox}

\paragraph{Vanilla mode (no tools)}

The no-tool baseline uses the system prompt below. Both prompts share
the task objective and image identifiers, but tool mode additionally provides
image sizes, tool and canvas conventions, and explicit planning and reflection
requirements. The comparison therefore measures the combined effect of
tool access and these prompt differences. A control that isolates tool
availability would need to match the image metadata and general reasoning
instructions across settings.

\begin{promptbox}
\ttfamily\footnotesize\raggedright\setlength{\parindent}{0pt}
\setlength{\parskip}{0pt}
You are an agent for multi-image visual reasoning. Keep working on the user's question until it is fully resolved, and only end your turn once you are sure of the answer. The question always has a correct answer, so never give up or claim it cannot be determined --- keep reasoning until you find it.
\par
\medskip
Solve the problem step by step. Look closely at the actual content of the images and reason about what they show.
\par
\medskip
The provided images are labelled in order as [img1], [img2], .... To point to a specific image, use exactly that imgN token.
\par
\end{promptbox}

\subsection{Training setup}
\label{sec:training_setup}
For each prompt, GRPO samples 8 rollouts and computes advantages using group-relative normalization without a learned critic. We use AdamW with a constant learning rate of $1 \times 10^{-6}$, a batch size of 32 prompts,  and 1 epoch. Both the KL coefficient and entropy coefficient are set to 0. Training is conducted on 4 NVIDIA H100 GPUs.

Rollouts are generated with a maximum prompt length of 16,384 tokens and a maximum response length of 20,480 tokens. Each multi-turn episode is limited to 10 assistant turns.
\section{\benchmarkname: A Dataset for Multi-Image Understanding}
\label{app:benchmark-details}
This section documents the construction of \benchmarkname{} and the associated training data.
It covers task definitions, source provenance, validation, and dataset construction procedures.

\subsection{Dataset Overview and Task Coverage}
\label{app:data-overview}

We construct \numtasks{} task types to produce \trainsetname{}
and \benchmarkname{}.
Table~\ref{tab:tasks} lists the tasks by their main visual challenge,
together with their source material and number of input images.
Source usage is documented in Sec.~\ref{app:data-sources},
and the selection of training and evaluation examples is described
in Sec.~\ref{sec:pipeline}.

{
\small
\setlength{\tabcolsep}{4pt}
\renewcommand{\arraystretch}{1.08}

\begin{longtable}{
    @{}
    >{\raggedright\arraybackslash}p{0.07\linewidth}
    >{\raggedright\arraybackslash}p{0.43\linewidth}
    >{\raggedright\arraybackslash}p{0.27\linewidth}
    >{\centering\arraybackslash}p{0.10\linewidth}
    @{}
}
\caption{
Task coverage of the \numtasks{}-type construction.
Tasks are grouped by their main visual challenge.
``Images'' denotes the number of input images per example.
}
\label{tab:tasks}\\

\toprule
\textbf{ID} &
\textbf{Task} &
\textbf{Source} &
\textbf{\# Images} \\
\midrule
\endfirsthead

\multicolumn{4}{@{}l}{\small\itshape Table~\ref{tab:tasks}, continued}\\
\toprule
\textbf{ID} &
\textbf{Task} &
\textbf{Source} &
\textbf{\# Images} \\
\midrule
\endhead

\midrule
\multicolumn{4}{r@{}}{\small\itshape Continued on the next page}\\
\endfoot

\bottomrule
\endlastfoot

\multicolumn{4}{@{}l}{\textbf{Resolution}}\\
\addlinespace[2pt]
1.1 & Small-target cross-frame tracking
    & VisDrone2019-MOT & 4 \\
1.3 & Distant sign reading
    & TT100K & 1 \\
1.5 & Pathology zoom
    & CAMELYON16 & 1 \\
1.6 & Micro-scratch detection
    & MVTec AD & 1 \\
1.7 & Dense small-object counting
    & CARPK & 1 \\

\addlinespace[5pt]
\multicolumn{4}{@{}l}{\textbf{Orientation}}\\
\addlinespace[2pt]
2.3 & Assembly orientation alignment
    & MVTec AD & 2 \\
2.5 & Mirror sign reading
    & TT100K & 1 \\
2.6 & Orthorectification
    & LEVIR-CD & 1 \\
2.8 & Oblique part rectification
    & MVTec AD & 2 \\
2.9 & Oblique document rectification
    & SmartDoc15-CH1 & 1 \\
2.10 & Text-orientation recovery
     & PubLayNet & 1 \\

\addlinespace[5pt]
\multicolumn{4}{@{}l}{\textbf{Precision comparison}}\\
\addlinespace[2pt]
3.8 & Bitemporal change detection
    & LEVIR-CD & 2 \\
3.10 & Golden-sample comparison
     & MVTec AD & 2 \\
3.11 & Symmetric self-comparison
     & MVTec AD & 1 \\
3.12 & Spot the difference
     & MSD & 2 \\
3.13 & GUI regression verification
     & Rico & 2 \\

\addlinespace[5pt]
\multicolumn{4}{@{}l}{\textbf{Hypothesis testing}}\\
\addlinespace[2pt]
4.2 & Camera-motion verification
    & VisDrone2019-MOT & 2 \\
4.3 & Template rotation for grasping
    & MVTec AD & 2 \\
4.6 & Multi-tile reassembly
    & LEVIR-CD & 4 \\
4.11 & Puzzle-piece restoration
     & COCO & 5 \\
4.12 & Mental rotation
     & Synthetic (polyominoes) & 5 \\

\addlinespace[5pt]
\multicolumn{4}{@{}l}{\textbf{Context interference}}\\
\addlinespace[2pt]
5.2 & Mirror reflection vs.\ real
    & MSD & 1 \\
5.5 & Simultaneous-contrast confusion
    & CAMELYON16 & 1 \\
5.6 & Illumination vs.\ defect
    & MVTec AD & 2 \\
5.7 & Illusion patch comparison
    & Synthetic (illusions) & 1 \\

\addlinespace[5pt]
\multicolumn{4}{@{}l}{\textbf{Spatial reference}}\\
\addlinespace[2pt]
6.1 & Trajectory grid coordinates
    & VisDrone2019-MOT & 4 \\
6.4 & Drivable-zone point query
    & BDD100K & 1 \\
6.5 & Change-coordinate localisation
    & LEVIR-CD & 2 \\
6.6 & Lesion-centre localisation
    & CAMELYON16 & 1 \\
6.7 & Defect coordinate report
    & MVTec AD & 1 \\
6.8 & Systematic grid scan
    & CARPK & 1 \\
6.9 & Zonal counting
    & CARPK & 1 \\

\end{longtable}
}

\subsection{Data Sources and Provenance}
\label{app:data-sources}

We use 12 public datasets and two procedural generators.
Table~\ref{tab:data-sources} lists the source material,
subsets used, and associated task IDs.

{
\small
\setlength{\tabcolsep}{4pt}
\renewcommand{\arraystretch}{1.12}
\begin{longtable}{
    @{}
    >{\raggedright\arraybackslash}p{0.15\linewidth}
    >{\raggedright\arraybackslash}p{0.30\linewidth}
    >{\raggedright\arraybackslash}p{0.29\linewidth}
    >{\raggedright\arraybackslash}p{0.20\linewidth}
    @{}
}
\caption{
Sources used to construct the task collection.
Task IDs follow Table~\ref{tab:tasks}.
Source subsets refer to the upstream datasets, not the final
training and evaluation sets.
}
\label{tab:data-sources}\\

\toprule
\textbf{Source Dataset} &
\textbf{Annotations} &
\textbf{Task IDs} \\
\midrule
\endfirsthead

\multicolumn{3}{@{}l}{
    \small\itshape Table~\ref{tab:data-sources}, continued
}\\
\toprule
\textbf{Source Dataset} &
\textbf{Annotations} &
\textbf{Task IDs} \\
\midrule
\endhead

\midrule
\multicolumn{3}{r@{}}{
    \small\itshape Continued on the next page
}\\
\endfoot

\bottomrule
\endlastfoot

VisDrone2019-MOT &
Aerial video frames and object-track annotations &
1.1, 4.2, 6.1 \\
\addlinespace[3pt]

TT100K &
Street images and traffic-sign annotations &
1.3, 2.5 \\
\addlinespace[3pt]

CAMELYON16 &
Whole-slide images and lesion annotations;
slide backgrounds for contrast tasks &
1.5, 5.5, 6.6 \\
\addlinespace[3pt]

MVTec AD &
Normal images, annotated defect images, and defect patches &
1.6, 2.3, 2.8, 3.10, 3.11, 4.3, 5.6, 6.7 \\
\addlinespace[3pt]

CARPK &
Aerial parking-lot images and car annotations &
1.7, 6.8, 6.9 \\
\addlinespace[3pt]

PubLayNet &
Document-page images &
2.10 \\
\addlinespace[3pt]

LEVIR-CD &
Co-registered aerial image pairs and building-change masks &
2.6, 3.8, 4.6, 6.5 \\
\addlinespace[3pt]

SmartDoc15-CH1 &
Document video frames and annotated page corners &
2.9 \\
\addlinespace[3pt]

MSD &
Scene images and mirror masks &
3.12, 5.2 \\
\addlinespace[3pt]

Rico &
Mobile screenshots and annotated UI-element boxes &
3.13 \\
\addlinespace[3pt]

COCO &
Natural-scene images &
4.11 \\
\addlinespace[3pt]

BDD100K &
Road images and drivable-area masks &
6.4 \\
\addlinespace[3pt]

Polyomino generator &
Chiral shapes rendered under rotations and reflections &
4.12 \\
\addlinespace[3pt]

Illusion generator &
Brightness-contrast, M\"uller-Lyer, and Ebbinghaus images &
5.7 \\

\end{longtable}
}

\paragraph{SmartDoc subset.}
SmartDoc15-CH1 has no official training split.
We use the first 24 of its 30 sorted document identifiers
across all five backgrounds, giving 120 videos.
The remaining six identifiers, comprising 30 videos,
are held out from construction.

\paragraph{Provenance records.}
Each sample records its source, source split, source fingerprint,
and source licence.
The \texttt{split\_exception} field records source-specific exceptions,
including the use of annotated MVTec AD defects, the custom SmartDoc
subset, and sources without an official split.
Source fingerprints support auditing of repeated source use.
The separation of training and evaluation examples at the source
level is described in Sec.~\ref{sec:pipeline}.

\paragraph{Overlap with external evaluation.}
We exclude an HPatches-based stitching-matrix verification task
because the same source corpus is used by a homography-estimation
task in an external multi-image benchmark.
The exclusion applies to the entire task type rather than to
individual samples.

\subsection{Sample Format and Shared Construction}
\label{app:data-format}

Each example is stored as a JSON record with the fields listed in
Table~\ref{tab:sample-schema}.

\begin{table}[htbp]
\centering
\caption{Shared sample schema.}
\label{tab:sample-schema}
\small
\setlength{\tabcolsep}{5pt}
\renewcommand{\arraystretch}{1.15}
\begin{tabular}{
    @{}
    >{\raggedright\arraybackslash}p{0.18\linewidth}
    >{\raggedright\arraybackslash}p{0.76\linewidth}
    @{}
}
\toprule
\textbf{Field} & \textbf{Content} \\
\midrule
\texttt{sample\_id} &
Sample identifier. \\
\texttt{prompt} &
Task question, input layout, and task-specific instructions. \\
\texttt{images} &
Input-image references in the order used by the prompt. \\
\texttt{task\_type} &
Task identifier corresponding to Table~\ref{tab:tasks}. \\
\texttt{target} &
Ground-truth answer used for scoring. \\
\texttt{metadata} &
Task-specific construction and audit information. \\
\bottomrule
\end{tabular}

\end{table}

\paragraph{Shared construction.}
Each task specifies an input selection or generation rule,
a question template, and a ground-truth relation.
Labels are derived from source annotations or known construction
parameters.
Task-specific filters check target visibility, spatial separation,
and geometric or assembly consistency.
The builders generate 400 examples for each of the \numtasks{}
task types, yielding 12,800 examples before training and evaluation
selection (Sec.~\ref{sec:pipeline}).
Validation procedures are described in Sec.~\ref{sec:validation},
and individual task constructions in Sec.~\ref{sec:tasks_template}.

\paragraph{Spatial conventions.}
Point and region coordinates are normalised and specified in the prompt.
Their pixel-coordinate equivalents are retained in \texttt{metadata}
for auditing and are not shown in the prompt.
For grid-based tasks, the prompt defines the grid size and
indexing convention.

\paragraph{Image resolution.}
Input images include source-resolution frames, level-0 whole-slide
crops, and views constructed on task-specific canvases.
Stored image dimensions and task-specific preprocessing are
specified in Sec.~\ref{sec:tasks_template}.

\paragraph{Scoring.}
The parsed final answer is evaluated by exact match against
\texttt{target.value}.
Missing or unparseable answers are scored as incorrect.

\subsection{Validation and Shortcut Checks}
\label{sec:validation}

We apply shared structural checks to all constructed examples
and validate labels against task-specific criteria.

\paragraph{Structural consistency.}
We check that the image list matches the declared input count,
every referenced file exists, and the target agrees with
its copy in the metadata.

\paragraph{Label validation.}
Region-based checks verify window bounds, overlap, and the
target-containment or coverage conditions specified by each task.
Point-localisation tasks enforce a tolerance around the ground-truth
position and a minimum separation from competing locations.
Homography and affine transformations are checked by corner
reprojection error, and rotations by circular angular error.
Assembly tasks use seam-continuity thresholds to distinguish the
original arrangement from alternative layouts or pieces.

Mask-based queries additionally check mirror coverage, overlap
with dilated masks, and window texture, or local drivable-zone
purity and membership.
The thresholds and additional checks for each task are specified
in Sec.~\ref{sec:tasks_template}.

\paragraph{Shortcut probes.}
We evaluate heuristics that use the supplied matrices, angles,
or coordinates without inspecting the images
(Table~\ref{tab:shortcut-probes}).
For camera-motion verification (4.2), candidate generation
uses rejection sampling against both the centroid and outlier
probes.

\begin{table}[t]
\centering
\caption{
Shortcut probes based on transformation and spatial geometry.
These heuristics do not inspect image content.
}
\label{tab:shortcut-probes}
\small
\setlength{\tabcolsep}{4pt}
\renewcommand{\arraystretch}{1.15}
\begin{tabular}{
    @{}
    >{\raggedright\arraybackslash}p{0.20\linewidth}
    >{\raggedright\arraybackslash}p{0.25\linewidth}
    >{\raggedright\arraybackslash}p{0.49\linewidth}
    @{}
}
\toprule
\textbf{Probe} & \textbf{Scope} & \textbf{Decision rule} \\
\midrule
Centroid / outlier &
Matrix-based tasks &
Select the transformation nearest to or furthest from
the mean of the supplied transformations. \\
\addlinespace[3pt]

Pair member / singleton &
Text-orientation recovery (2.10) &
Select an angle from the pair separated by $180^\circ$,
or the angle outside that pair. \\
\addlinespace[3pt]

Central-$x$ / lowest-$y$ &
Drivable-zone point query (6.4) &
Select the most horizontally central point or the lowest
point in the image. \\
\bottomrule
\end{tabular}

\end{table}

\paragraph{Source-annotation audit.}
For a subset of examples, ground-truth labels are re-derived
directly from source annotations rather than from the builder's
intermediate state.
The audit results are logged with the build.

\subsection{Training and Evaluation Set Construction}
\label{sec:pipeline}
The candidate pool is filtered using tool-enabled and no-tool rollouts to form the training suite.
The evaluation set is constructed with overlap checks against the selected training data.

\paragraph{Rollout-based selection.}
For each constructed example, we run the same policy eight times
with tool access and once without tools.
The tool-enabled runs use temperature sampling with distinct seeds.
Question prompts, answer parsing, and scoring are held fixed
across the two settings.
Let $k\in\{0,\ldots,8\}$ denote the number of correct tool-enabled
runs.
Table~\ref{tab:pipeline} summarises the outcomes and the resulting
selection.

\paragraph{Selecting \trainsetname{}.}
We retain examples answered incorrectly in the no-tool run
and correctly in two to seven of the eight tool-enabled runs.
The resulting training suite contains 3,509 examples and
7,155 images across 32 task types.
Orthorectification (2.6) contributes no examples because all
its success counts are either $k=0$ or $k=8$.

\paragraph{Held-out hard subset.}
The 4,032 examples answered incorrectly without tools and correctly
in at most one tool-enabled run are retained as a hard subset.
They span all \numtasks{} task types and are not used for training.

\begin{table}[t]
\centering
\caption{
Rollout outcomes and data selection.
The second row contains examples answered incorrectly in the
single no-tool run.
The final three rows partition the constructed examples.
}
\label{tab:pipeline}
\small
\setlength{\tabcolsep}{3pt}
\renewcommand{\arraystretch}{1.1}
\begin{tabular}{@{}l*{10}{r}@{}}
\toprule
& \multicolumn{9}{c}{\textbf{Tool successes $k$ out of eight runs}}
& \\
\cmidrule(lr){2-10}
& 0 & 1 & 2 & 3 & 4 & 5 & 6 & 7 & 8
& \textbf{Total} \\
\midrule
Constructed examples
& 3116 & 1647 & 1268 & 995 & 769 & 756 & 715 & 930 & 2604
& 12{,}800 \\
Incorrect without tools
& 2603 & 1429 & 1026 & 735 & 546 & 449 & 395 & 358 & 669
& 8{,}210 \\
\midrule
\trainsetname{}
& -- & -- & 1026 & 735 & 546 & 449 & 395 & 358 & --
& 3{,}509 \\
Held-out hard subset
& 2603 & 1429 & -- & -- & -- & -- & -- & -- & --
& 4{,}032 \\
Remaining examples
& 513 & 218 & 242 & 260 & 223 & 307 & 320 & 572 & 2604
& 5{,}259 \\
\bottomrule
\end{tabular}

\end{table}

\paragraph{Constructing \benchmarkname{}.}
We select benchmark examples from the constructed data after
removing overlap with \trainsetname{} at three levels:
\begin{enumerate}
    \item \textit{Sample identity.}
    Exclude examples whose \texttt{sample\_id} appears in training.

    \item \textit{Source provenance.}
    Exclude examples sharing a source container with any training
    example, across all task types.
    The source unit is a VisDrone scene, a CAMELYON slide,
    a SmartDoc video, or otherwise an individual source image.

    \item \textit{Image similarity.}
    Exclude examples containing an image byte-identical to a
    training image.
    For synthetic and CARPK tasks, every retained image must
    additionally have a distance of at least four from every
    training image under a 64-bit perceptual hash.
\end{enumerate}

The image-level checks address duplicate synthetic renders
and near-identical frames with different source identifiers.
Visually similar product and document images from distinct
physical instances are retained and listed in the benchmark manifest.

The resulting \benchmarkname{} contains 560 examples across
28 task types, with 20 examples per type.
The three VisDrone-derived tasks (1.1, 4.2, and 6.1) are excluded
because training examples cover all 56 source sequences.
Mental rotation (4.12) is excluded because no synthetic render
survives the image-level filter.

\subsection{Task-Specific Construction}
\label{sec:tasks_template}

For each task, we describe the inputs, question, ground-truth
relation, and task-specific construction checks.
Shared sample conventions and validation procedures are described
in Secs.~\ref{app:data-format} and~\ref{sec:validation}.

\subsubsection{Resolution}

These tasks involve small visual targets or dense collections
of objects.
Target-size records include the extent relative to the source
frame and the corresponding size at a $768$-pixel reference
input scale.

\paragraph{1.1 Small-target cross-frame tracking.}
\emph{Inputs and construction.}
Four $1904\times1071$ frames are sampled from one VisDrone2019-MOT
sequence, with a per-example stride of 10--35 frames.
One annotated track is designated as the target.
Its mean width is $1.9\%$ of the frame width, corresponding to
approximately 15 pixels at the reference input scale.

\emph{Question.}
Given the target's normalised position in the first frame,
identify its position in the fourth frame.

\emph{Ground truth and checks.}
The target position is obtained from the track annotation.
A correct candidate lies within a normalised Chebyshev distance
of $0.02$ from this position; distractors are at least $0.08$ away.
Distractors use the fourth-frame positions of other annotated
tracks in the same scene.
In negative examples, no candidate lies within $0.08$ of
the target position.

\paragraph{1.3 Distant sign reading.}
\emph{Inputs and construction.}
A $2048\times2048$ TT100K street image contains exactly one
speed-limit sign.
The sign's longest side is restricted to 22--44 pixels
in the source image.

\emph{Question.}
Read the speed limit displayed on the sign.

\emph{Ground truth and checks.}
The answer is derived from the sign's annotated class.
Distractor values are drawn from other speed limits in
the same sign family.

\paragraph{1.5 Pathology zoom.}
\emph{Inputs and construction.}
A $2048\times2048$ region is read at level~0 from a CAMELYON16
whole-slide image.
The region contains an annotated metastatic lesion a few
hundred level-0 pixels across.

\emph{Question.}
Identify which proposed region contains metastatic tumour.

\emph{Ground truth and checks.}
Labels are derived from the lesion annotation.
A positive window fully contains the lesion, which touches
no other candidate window.
All candidate windows have equal dimensions and satisfy
a tissue-coverage threshold.

\paragraph{1.6 Micro-scratch detection.}
\emph{Inputs and construction.}
A $1024\times1024$ MVTec AD image contains a real annotated
surface defect, such as a scratch, cut, crack, poke, or thread.
Candidate windows have equal dimensions and a width of
approximately $0.12$ of the image width.

\emph{Question.}
Identify the window containing the defect.

\emph{Ground truth and checks.}
Labels are derived from the defect mask.
A positive window fully contains the mask, which touches
no distractor window.
All candidate windows lie on the part surface.

\paragraph{1.7 Dense small-object counting.}
\emph{Inputs and construction.}
A $1280\times720$ CARPK parking-lot image is paired with
complete car annotations.
Per-quadrant counts are recorded and used during sampling
to vary the spatial distribution of cars.

\emph{Question.}
Count all cars in the image.

\emph{Ground truth and checks.}
The answer is the total annotated car count.
Distractors are nearby counts, and the rank of the true count
among the numerical candidates varies across examples.

\subsubsection{Orientation}

These tasks involve recovering or identifying image geometry
under rotation, reflection, and perspective changes.

\paragraph{2.3 Assembly orientation alignment.}
\emph{Inputs and construction.}
A normal MVTec AD image provides the reference orientation.
A known rotation produces a second view of the same part.

\emph{Question.}
Identify the counterclockwise rotation that aligns the query
view with the reference.

\emph{Ground truth and checks.}
The target angle is determined by the construction rotation.
Candidate angles are separated by at least $30^\circ$.
The transformed views are checked for visual distinguishability
to exclude ambiguous cases caused by approximate rotational symmetry.

\paragraph{2.5 Mirror sign reading.}
\emph{Inputs and construction.}
A crop containing a TT100K speed-limit sign is reflected
horizontally.

\emph{Question.}
Read the speed limit displayed on the reflected sign.

\emph{Ground truth and checks.}
The label is the annotated class of the original sign.

\paragraph{2.6 Orthorectification.}
\emph{Inputs and construction.}
A $1024\times1024$ LEVIR-CD aerial image is warped by a known
projective transformation to produce an oblique view.

\emph{Question.}
Identify the homography that reverses the perspective distortion.

\emph{Ground truth and checks.}
The ground-truth matrix is the inverse of the applied warp,
with zero corner reprojection error normalised by the image diagonal.
Distractor matrices are perturbations whose errors fall within
a task-specific non-zero band.
In negative examples, the inverse is omitted and every proposed
matrix exceeds the specified error floor.

\paragraph{2.8 Oblique part rectification.}
\emph{Inputs and construction.}
An MVTec AD part image is warped by a known homography.
The original image is supplied alongside the warped view
as the rectification reference.

\emph{Question.}
Identify the homography that maps the oblique view
back to the reference.

\emph{Ground truth and checks.}
The ground-truth matrix inverts the applied warp.
Corner-error checks and distractor construction follow Task~2.6.

\paragraph{2.9 Oblique document rectification.}
\emph{Inputs and construction.}
A $1920\times1080$ frame is selected from SmartDoc15-CH1.
Annotated document corners define the mapping to an upright
page with A4 aspect ratio.

\emph{Question.}
Identify the homography that rectifies the document
to the target page geometry.

\emph{Ground truth and checks.}
The ground-truth homography maps the annotated quadrilateral
to the upright page.
Corner-error checks follow Task~2.6.

\paragraph{2.10 Text-orientation recovery.}
\emph{Inputs and construction.}
A PubLayNet document page is rotated by a known angle.
Pages are selected to contain enough text lines to define
an upright reading orientation.

\emph{Question.}
Identify the additional counterclockwise rotation that restores
upright text.

\emph{Ground truth and checks.}
A correct restoring angle has zero circular error relative
to the construction-derived angle.
Incorrect angles differ by at least $25^\circ$.
Each example includes exactly one pair of proposed angles
separated by $180^\circ$.
The corresponding angle-structure probes are described in
Sec.~\ref{sec:validation}.

\subsubsection{Precision comparison}

These tasks concern local changes or asymmetries.
Selected constructions introduce photometric or encoding variation
alongside the target difference.

\paragraph{3.8 Bitemporal change detection.}
\emph{Inputs and construction.}
Two co-registered $1024\times1024$ LEVIR-CD images depict
the same area at different dates.
Building-change masks distinguish the target changes from
other appearance variation.

\emph{Question.}
Locate building construction or demolition between the two images.

\emph{Ground truth and checks.}
A positive window contains at least the task-specific minimum
number of annotated change pixels.
The change mask does not intersect any distractor window.
Candidate windows have equal dimensions.

\paragraph{3.10 Golden-sample comparison.}
\emph{Inputs and construction.}
A normal MVTec AD image serves as the reference.
A pixel-aligned copy receives a real annotated defect patch
in positive examples.
Independent photometric jitter is then applied to both views.

\emph{Question.}
Identify the region containing a defect absent from the reference.

\emph{Ground truth and checks.}
The composited defect mask determines the target window.
Photometric differences outside the mask are unrelated
to the defect label.

\paragraph{3.11 Symmetric self-comparison.}
\emph{Inputs and construction.}
An MVTec AD image is made bilaterally symmetric by mirroring
one half onto the other.
In positive examples, a real defect patch is composited
onto one side.

\emph{Question.}
Identify the region containing the defect that breaks
the expected symmetry.

\emph{Ground truth and checks.}
The composited mask identifies the defective region.
Its mirrored counterpart is included as a distractor
in every positive example.

\paragraph{3.12 Spot the difference.}
\emph{Inputs and construction.}
Two copies of an MSD scene are encoded at JPEG quality levels
92 and 88.
Positive examples contain one synthetic local edit;
negative examples contain only the encoding differences.

\emph{Question.}
Locate the edited region, distinguishing it from compression noise.

\emph{Ground truth and checks.}
The recorded edit location determines the target window.
Examples without an edit are labelled as having no target change.

\paragraph{3.13 GUI regression verification.}
\emph{Inputs and construction.}
Two $1080\times1920$ views are derived from a Rico screenshot
and encoded at different JPEG qualities.
In positive examples, one annotated UI element is recoloured,
shifted by a few pixels, or removed.
The edit magnitude is recorded.

\emph{Question.}
Identify the region containing the changed UI element.

\emph{Ground truth and checks.}
Candidate regions are based on annotated UI-element boxes.
The target region contains the box of the edited element.

\subsubsection{Hypothesis testing}

These tasks evaluate proposed transformations, arrangements,
or shape matches against visual evidence.

\paragraph{4.2 Camera-motion verification.}
\emph{Inputs and construction.}
Two $768\times768$ views are derived from a VisDrone frame.
The second is generated from the first by a known $2\times3$
affine transformation comprising translation, scaling, and rotation.

\emph{Question.}
Identify the affine transformation that maps the first view
to the second.

\emph{Ground truth and checks.}
The construction matrix has zero corner reprojection error.
Distractor matrices lie within a specified non-zero error band.
Candidate generation uses rejection sampling against the centroid
and outlier probes described in Sec.~\ref{sec:validation}.

\paragraph{4.3 Template rotation for grasping.}
\emph{Inputs and construction.}
An MVTec AD image serves as a canonical template.
A known rotation generates the observed view.

\emph{Question.}
Identify the counterclockwise rotation that maps the template
to the observed view.

\emph{Ground truth and checks.}
The target is the applied rotation angle.
Candidate angles are separated by at least $30^\circ$,
and the resulting views are checked for visual distinguishability.

\paragraph{4.6 Multi-tile reassembly.}
\emph{Inputs and construction.}
Four $512\times512$ tiles are cut from a LEVIR-CD image
on a $2\times2$ grid and presented in shuffled order.

\emph{Question.}
Identify the assignment of tiles to grid positions that restores
the original image.

\emph{Ground truth and checks.}
The original tile-to-position assignment defines the label.
Its mean absolute seam discontinuity must fall below
a task-specific threshold.
Every distractor arrangement must exceed a separate,
higher threshold.

\paragraph{4.11 Puzzle-piece restoration.}
\emph{Inputs and construction.}
The inputs comprise a $480\times480$ COCO image with a square
region removed and candidate pieces shown under sampled rotations.
The candidates include the original piece, its reflection,
and patches from other locations in the same image.

\emph{Question.}
Identify the piece that restores the missing region.

\emph{Ground truth and checks.}
The original piece must produce a seam error below the acceptance
threshold when restored to its original position and orientation.
Each distractor must exceed a higher error threshold.

\paragraph{4.12 Mental rotation.}
\emph{Inputs and construction.}
Five $448\times448$ images depict a chiral polyomino:
a reference, a rotated copy, and reflected copies
at different orientations.

\emph{Question.}
Identify the figure related to the reference by an in-plane
rotation without reflection.

\emph{Ground truth and checks.}
Labels follow the known rotation and reflection operations
used to generate each image.
The rotated copy is the target; reflected copies are distractors.

\subsubsection{Context interference}

These tasks distinguish target properties from surrounding
structure, reflections, or photometric variation.

\paragraph{5.2 Mirror reflection vs.\ real.}
\emph{Inputs and construction.}
An MSD indoor image contains an annotated mirror.
Equal-sized candidate windows are selected inside and outside
the mirror region.

\emph{Question.}
Identify a window containing only mirror reflection.

\emph{Ground truth and checks.}
The positive window must have at least $98\%$ mirror coverage
under both the original mask and its dilated version.
Distractor windows have zero overlap with the dilated mask.
Every window must meet a minimum greyscale standard deviation
to exclude untextured regions.

\paragraph{5.5 Simultaneous-contrast confusion.}
\emph{Inputs and construction.}
A low-magnification CAMELYON16 slide image provides the background
for three uniform grey patches with mid-tone, bright,
and dark surrounds.
Positive examples change one patch by a signed intensity
difference of magnitude 6--12 on the 0--255 scale.
Other examples retain identical patch intensities.

\emph{Question.}
Determine whether one patch differs in grey value
and identify it when present.

\emph{Ground truth and checks.}
Labels are computed from the printed patch intensities,
independently of their surrounding backgrounds.

\paragraph{5.6 Illumination vs.\ defect.}
\emph{Inputs and construction.}
Two pixel-aligned MVTec AD views depict the same part.
The second receives synthetic spotlight illumination.
Positive examples additionally contain a composited real defect;
negative examples contain only the lighting change.
The illumination perturbation produces larger raw intensity
differences than the defect.

\emph{Question.}
Locate a real defect while disregarding the lighting change.

\emph{Ground truth and checks.}
The composited defect mask determines the target region.
Relighting-only examples are labelled as having no defect.

\paragraph{5.7 Illusion patch comparison.}
\emph{Inputs and construction.}
Images are generated from three illusion families:
M\"uller-Lyer figures with horizontal segments and terminal fins,
Ebbinghaus figures with central and surrounding circles,
and brightness-contrast figures with squares on different backgrounds.
The compared elements have known pixel lengths, diameters,
or intensities.

\emph{Question.}
Compare the designated elements by length, diameter,
or intensity, including the case of equality.

\emph{Ground truth and checks.}
The relation is computed directly from the rendering parameters.
Each record also stores the relation suggested by
the surrounding illusion.

\subsubsection{Spatial reference}

These tasks associate visual evidence with a specified
coordinate system, grid cell, or region.

\paragraph{6.1 Trajectory grid coordinates.}
\emph{Inputs and construction.}
Four VisDrone2019-MOT frames from a fixed viewpoint contain
an annotated target track.
The prompt defines a $6\times6$ grid with 1-based
row and column indexing.

\emph{Question.}
Report the target's grid-cell sequence across the four frames.

\emph{Ground truth and checks.}
The sequence is computed from the annotated track positions.
Distractors follow other real tracks in the same scene
and differ from the target sequence in at least two frames.

\paragraph{6.4 Drivable-zone point query.}
\emph{Inputs and construction.}
A $1280\times720$ BDD100K road image is selected with
drivable-area annotations distinguishing direct, alternative,
and non-drivable regions.

\emph{Question.}
Identify a normalised point on the directly drivable corridor
ahead of the ego vehicle.

\emph{Ground truth and checks.}
The positive point lies in the directly drivable region;
distractors lie outside it.
Each point's local patch must have at least $95\%$ zone purity,
and candidate points have a minimum separation of $0.08$.
The central-$x$ and lowest-$y$ probes are described in
Sec.~\ref{sec:validation}.

\paragraph{6.5 Change-coordinate localisation.}
\emph{Inputs and construction.}
A co-registered LEVIR-CD image pair is selected with
its annotated building-change mask.

\emph{Question.}
Report the centre of the main building change
as a normalised $(x,y)$ point.

\emph{Ground truth and checks.}
The target is derived from the change-mask centroid.
The correct point lies within $0.07$ of the centroid,
while distractors are at least $0.2$ away.

\paragraph{6.6 Lesion-centre localisation.}
\emph{Inputs and construction.}
A level-0 CAMELYON16 slide region contains exactly one
annotated metastatic lesion.

\emph{Question.}
Report the lesion centre as a normalised point.

\emph{Ground truth and checks.}
The lesion annotation defines the target centre.
The localisation tolerance is $0.07$, with a minimum
distractor separation of $0.2$.

\paragraph{6.7 Defect coordinate report.}
\emph{Inputs and construction.}
An MVTec AD image contains exactly one annotated defect region.

\emph{Question.}
Report the defect centre as a normalised point.

\emph{Ground truth and checks.}
The defect annotation defines the target centre.
The localisation tolerance is $0.06$, with a minimum
distractor separation of $0.18$.

\paragraph{6.8 Systematic grid scan.}
\emph{Inputs and construction.}
A CARPK image with complete car annotations is paired with
a grid whose size varies from $4\times4$ to $8\times8$.
The prompt specifies 1-based row and column indexing.

\emph{Question.}
Identify the unique cell containing exactly $k$ car centres,
where $k\in\{1,2\}$.

\emph{Ground truth and checks.}
Annotation-derived counts verify that exactly one cell
satisfies the query.
Distractor cells are empty or contain at least four cars.
An edge guard keeps car centres away from cell boundaries.

\paragraph{6.9 Zonal counting.}
\emph{Inputs and construction.}
A CARPK image is paired with a grid ranging from
$5\times5$ to $8\times8$, with one cell designated as the query.
Grid indexing follows Task~6.8.

\emph{Question.}
Count the cars whose centres lie in the specified cell.

\emph{Ground truth and checks.}
The answer is computed from the annotated car centres.
Boundary checks follow Task~6.8.
The rank of the true count among the proposed numerical values
varies across examples.

\subsection{Reproducibility and Release Details}
\label{app:data-reproducibility}
This section records construction seeds, source access procedures, and release information
needed to reproduce the task collection.

\paragraph{Construction seeds.}
The builders use seed \texttt{20260807} for 32 task types
and \texttt{20260808} for the remaining type.
Seeded sampling controls source selection, window placement,
and distractor generation.

\paragraph{Source access.}
The builders access source data without extracting complete archives.
MVTec AD is read from a local ZIP archive.
CAMELYON16, TT100K, and COCO are accessed through HTTP range
requests that retrieve the selected source content.
LEVIR-CD and Rico are read row-wise from Parquet files.
These access patterns avoid requiring full local copies
of the remote source files.

\paragraph{Source attribution.}
Source attribution and licence information are recorded
per sample, as described in Sec.~\ref{app:data-sources}.

\clearpage
\section{Extended Results}
\label{app:extended-results}
This section reports additional tool and model comparisons and examines failure profiles.
These analyses complement the main results by characterising tool use and intermediate answers.

\subsection{Beyond Cropping}
We evaluate the same trained \modelname{} checkpoint with
either crop alone or the full visual toolset
(Table~\ref{tab:crop-only-ablation}).
The full toolset increases accuracy by 9.4 percentage points
on \benchmarkname{} and 3.0 points on M4Bench.
Gains are smaller on Mantis and BLINK, at 1.0 and 0.4 points,
respectively, while accuracy on MMIU decreases by 1.1 points.
The benefit of operations beyond cropping is therefore
most pronounced on \benchmarkname{}.

\begin{table}[h]
\centering
\caption{\textbf{Visual tool ablation of Mosaic Agent.}
We compare the full visual toolset with a restricted setting
where only the Crop tool is available.}
\label{tab:crop-only-ablation}
\small
\setlength{\tabcolsep}{7.0pt}
\renewcommand{\arraystretch}{1.10}
\begin{tabular}{@{}lccccc@{}}
\toprule
\textbf{Available Visual Tools}
& \textbf{MosaicBench}
& \textbf{Mantis}
& \textbf{BLINK}
& \textbf{M4Bench}
& \textbf{MMIU}
\\

\midrule
Crop only
& .539
& .806
& .624
& .588
& .583
\\

Full toolset
& .633{\tiny$\pm$.011}
& .816{\tiny$\pm$.012}
& .628{\tiny$\pm$.010}
& .618{\tiny$\pm$.014}
& .572
\\

\bottomrule
\end{tabular}

\end{table}

\begin{table}[H]
\centering
\caption{\textbf{Comparison between Qwen3-VL-8B and Mosaic Agent}
under matched inference settings.}
\label{tab:base-vs-agent}

\small
\setlength{\tabcolsep}{5.5pt}
\renewcommand{\arraystretch}{1.08}

\begin{tabular}{@{}llccccc@{}}
\toprule

\textbf{Model}
& \textbf{Setting}
& \textbf{MosaicBench}
& \textbf{Mantis}
& \textbf{BLINK}
& \textbf{M4Bench}
& \textbf{MMIU}
\\

\midrule

\multirow{2}{*}{Qwen3-VL-8B}
& \TRe{}
& .452{\tiny$\pm$.043}
& .819{\tiny$\pm$.010}
& .644{\tiny$\pm$.004}
& .513{\tiny$\pm$.012}
& .610
\\

& \VRe{}
& .472{\tiny$\pm$.015}
& .797{\tiny$\pm$.016}
& .636{\tiny$\pm$.002}
& .518{\tiny$\pm$.003}
& .566
\\

\midrule

\multirow{2}{*}{\textbf{\modelname{}}}
& \TRe{}
& .460{\tiny$\pm$.012}
& .828{\tiny$\pm$.004}
& .637{\tiny$\pm$.004}
& .513{\tiny$\pm$.014}
& .594
\\

& \VRe{}
& .633{\tiny$\pm$.011}
& .816{\tiny$\pm$.012}
& .628{\tiny$\pm$.010}
& .618{\tiny$\pm$.014}
& .572
\\

\bottomrule
\end{tabular}
\end{table}

\FloatBarrier

\noindent
\begin{minipage}{\linewidth}
\subsection{Failure Patterns}
The profiles in Fig.~\ref{fig:voi_pattern} capture three failure behaviours:
losing an initially correct probe answer, obtaining a correct intermediate
probe answer but later losing it, and continued processing without a
correct probe answer. These profiles describe the timing of failure and
motivate examining answer retention and stopping decisions.

\medskip
    \centering
    \includegraphics[width=1\linewidth]{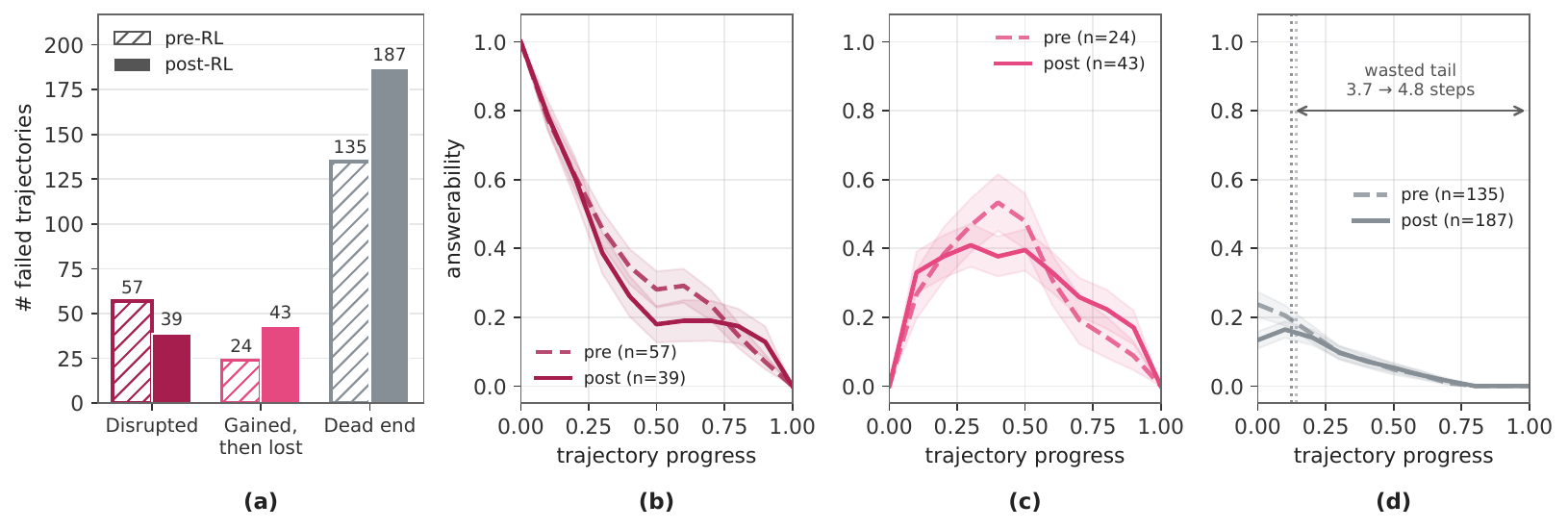}
\captionof{figure}{\textbf{Answerability profiles before and after RL on failed tasks.}
Training leaves the number of failures almost unchanged ($231\rightarrow221$)
but reshapes them: fewer trajectories destroy an answer the model already had
(a, b), while more find an answer they cannot hold (a, c) and more settle into
a wrong answer several steps before stopping (a, d).}
\label{fig:voi_pattern}
\end{minipage}
\par

\end{document}